\documentclass[conference]{IEEEtran}
\IEEEoverridecommandlockouts
\usepackage{cite}
\usepackage{amsmath,amssymb,amsfonts}
\usepackage{algorithmic}
\usepackage{graphicx}
\usepackage{textcomp}
\usepackage{xcolor}
\usepackage{hyperref}
\hypersetup{
    colorlinks=true,
    linkcolor=blue,
    filecolor=magenta,      
    urlcolor=cyan,
    citecolor=black,
    pdftitle={Sharelatex Example},
    bookmarks=true,
    pdfpagemode=FullScreen,
    }
\def\BibTeX{{\rm B\kern-.05em{\sc i\kern-.025em b}\kern-.08em
    T\kern-.1667em\lower.7ex\hbox{E}\kern-.125emX}}
\begin{document}

% \title{Keep competing, from home\\
% \title {Keep competing, from home: Automatic Scoring Benchmark for a Virtual Environment Competition Testbed}
\title{Test Framework for a Virtual Competition Testbed}
% {\footnotesize \textsuperscript{*}Note: Sub-titles are not captured in Xplore and
% should not be used}
% }

% proper format is really long,found the following alternative:
\author{\IEEEauthorblockN{Liam Wellacott\IEEEauthorrefmark{1},
Emilyann Nault\IEEEauthorrefmark{2}, Ioannis Skottis\IEEEauthorrefmark{3}
***Alexandre Colle\IEEEauthorrefmark{4}, Shreyank N Gowda\IEEEauthorrefmark{5}, \\Pierre Nicolay\IEEEauthorrefmark{6}, and Emily Rolley-Parnell\IEEEauthorrefmark{7}}
\IEEEauthorblockA{Heriot-Watt University \& University of Edinburgh\\
Edinburgh, United Kingdom\\
Email: \IEEEauthorrefmark{1}lw88@hw.ac.uk,
\IEEEauthorrefmark{2}en27@hw.ac.uk,
\IEEEauthorrefmark{3}i.skottis@ed.ac.uk,
\IEEEauthorrefmark{4}ac385@hw.ac.uk,
\IEEEauthorrefmark{5}s1960707@ed.ac.uk,\\
\IEEEauthorrefmark{6}pon1@hw.ac.uk,
\IEEEauthorrefmark{7}emily.rolley-parnell@ed.ac.uk}
***The following authors are ordered alphebetically\\
Code: \url{https://github.com/LiamWellacott/CDT2019-ERL}\\
Video: \url{https://www.youtube.com/watch?v=RoxGla2P-uY&ab_channel=ColleAlexandre}
}

\maketitle

% Guidance from UK RAS:
% Authors are encouraged to address the following key points in their paper: 

% Purpose 
% Originality and value 
% Design, methodology and approach 
% Results and findings (if paper is empirical) N/A
% Research limitations and implications 
% In addition, you may provide up to six keywords that encapsulate the principal topics covered and help categorise your paper

% Challenges working from home also encouraged

\begin{abstract}

Virtual environments have been utilised in robotics research as a tool to assess systems before deploying them in the field. The COVID-19 pandemic has brought about additional motivation for the development of virtual benchmarks in order to aid in safe and productive development. In-person robotics competitions have also halted, thus limiting the scope of opportunities for students and researchers. We implemented the structure of a service robotics competition into an extendable and adaptable virtual scoring environment. The competition challenges the state of the art in home service robotics by presenting realistic household tasks for robots to complete. The virtual environment provides a foundation for competition teams to assess their systems when accessing the physical environment is not possible. We believe that utilising virtual environments as a means of assessment will lead to other benefits such as increased access and generalisation.

\end{abstract}


\section{Introduction}
% Purpose 
Competitions are an important tool for pushing the state of the art in robotics applications. They also provide new researchers with practical skills and knowledge used throughout their career. Hosting physical competitions has not been possible during the COVID-19 pandemic. This has highlighted the need for tools which allow teams to continue their work regardless of access to the physical competition environment. 

%\cite{lima2016} competition
% \cite{ERLHandbook}
In this work we target the \href{https://www.eu-robotics.net/robotics_league/}{European Robotics League Consumer Service Robots} competition (ERL Consumer for short). Teams are challenged with performing realistic household tasks for an older adult, Granny Annie. The competition is normally hosted at locations across Europe with accredited testbeds designed to appear like a typical small apartment. The competition organisers provide a \href{https://www.eu-robotics.net/robotics_league/upload/documents-2018/ERL_Consumer_10092018.pdf}{handbook} which describes the tasks and scoring criteria. Additionally, functional benchmarks provide a method to assess a team's solution in a single capability (e.g. natural language understanding). Both the tasks and functional benchmarks are performed across a three-day period. They are partially automated, but require human intervention for purposes such as setting up objects in the environment. 

To our knowledge, there is no automated virtual environment where teams can test their system against the ERL competition tasks and functional benchmarks. We identified this gap first-hand when attempting to produce our own competition entry while working under the restrictions of the COVID-19 lockdown in the UK. In this paper we present our virtual benchmark for the ERL Consumer competition.

\section{Background}

% In robotics research, virtual environments have been created for a variety of reasons including aiding in data collection \cite{rossmann2013} and testing. In the field of human-robot interaction (HRI), these environments are limited when it comes to robot embodiment \cite{wainer2007}. However, virtual environments have numerous benefits such as  increasing access and safety considerations \cite{rossmann2013}. Physical platforms (robots and testbeds) are expensive to run and maintain, and in-development features risk damaging expensive equipment \cite{AutonomousDrivingSimulator2017}. Researchers have thus transitioned into hybrid approaches when developing new algorithms or strategies for robotics \cite{MobileRobotsSim2019, PepperRobotSimulation2018}. In the time of the pandemic, tools such as the Robotic Programming Network \cite{casan2015} can provide a means for researchers and students to interact with the physical robot remotely.

%\cite{NASASRC} space robotics
% \cite{RoboCup@Home}
% \cite{RoboCupRescue}
Robotics competitions have made a shift to becoming more virtual, which has only increased in prominence since the start of the COVID-19 pandemic. Some competitions such as the \href{www.spaceroboticschallenge.com}{SpaceRobotsChallenge} have opted to use simulation since sending competition robots to space would be infeasible/unreasonable. In recognition of the disruptive nature of the COVID-19 pandemic, other competitions have shifted towards virtual platforms such the VirtualRobotX competition. More recently other competitions have announced virtual competitions entries such as 2021 \href{https://athome.robocup.org/}{RoboCup@Home} and 2021 \href{https://www.robocup.org/domains/2}{RoboCup Rescue}.

In the context of service robotics, OpenRobotics, in collaboration with Hitatchi, created a virtual robotics competition called ServiceSim\cite{ServiceSim}. The competition's focus is HRI in an office environment. In an effort to facilitate research in the field of social robotics and specifically the learning of social norms (e.g. proxemics), Pimentel and Aquino-Junior \cite{SocialRulesSimulationOpenai2020} have developed a virtual environment with a scoring mechanism using ROS, Gazebo and OpenAI. Most recently, SEAN by Nathan Tsoi \textit{et al.} \cite{tsoi2020sean} developed a virtual world with the focus on the development and evaluation of algorithms for social navigation (dynamic conditions such as simulated pedestrians, cars, etc.). 

Robotics competitions are not only important for robotics research, but also are used to educate and engage students.  %However, the cost of robotics platforms has kept robotics out of reach for many teaching environments \cite{EducationSimulatorReview2021}. This has resulted in a push for students to build inexpensive systems. 
A good example of this is the competition RoboCupJunior \cite{eguchi2016}. % which had an entry utilising a PlayStation Eye webcam and a RaspberryPi for image processing.
Virtual environments can grant students access to state of the art equipment in cases where it otherwise would not have been possible.

% Additionally, Yanco \emph{et al.} \cite{yanco2004} produced HRI development guidelines after reviewing four entries from the American Association for Artificial Intelligence Robot Rescue Competition.

\section{Virtual Benchmark}
% Design, methodology and approach
% describe what we have

The virtual benchmark is a system for testing ROS-based ERL Consumer competition solutions in an automated virtual environment. When the user starts the benchmark, the \textit{simulation environment}, which has been configured for a given \textit{scenario}, is launched. The user's competition solution performs the task and notifies the \textit{virtual referee} upon completion. The referee evaluates the task performance and returns a score.

\begin{figure}[htbp]
\centerline{\includegraphics[width=7cm]{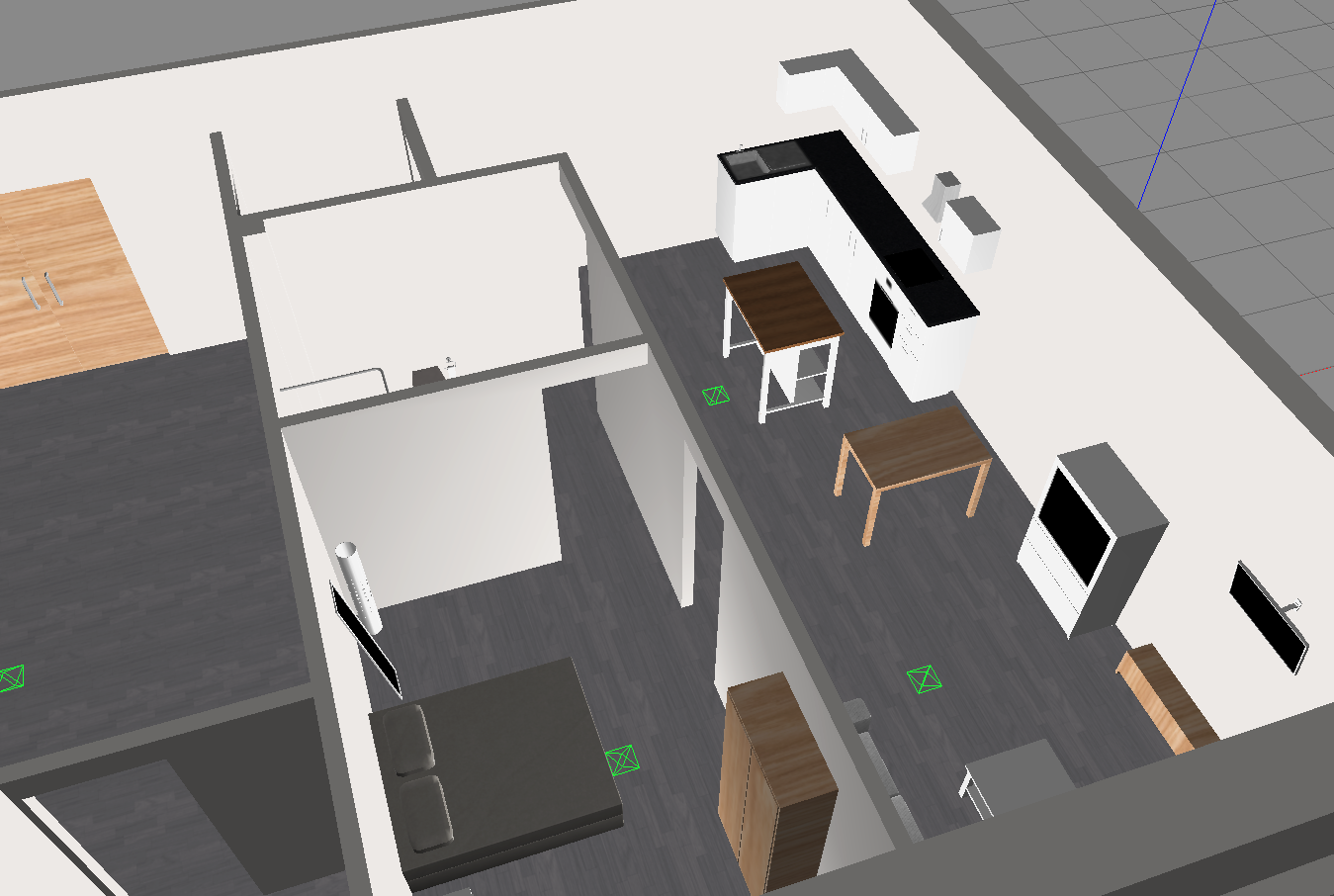}}
\caption{The virtual RALT Gazebo world}
\label{virtualralt}
\end{figure}

% \cite{RALT}
The \textit{simulation environment} is a Gazebo world designed with the same furniture and overall dimensions as the physical Heriot-Watt University \href{https://ralt.hw.ac.uk/}{Robotic Assisted Living Testbed (RALT)} (Fig. \ref{virtualralt}). To configure the world for a particular scenario, we utilise the large number of Gazebo-compatible object models available online (e.g. boxes of food, standing human). Each scenario is built using a separate \texttt{roslaunch} file which describes the positions of the objects as well as the starting position of the robot. As in the physical competition, the team has time to ensure the robot is fully initialised before the scenario begins.    

To start the scenario, the user launches the \textit{virtual referee}, which is based on the Referee, Scoring and Benchmarking Box (RSBB) used in the ERL Consumer competition. In addition to its role in scoring and scenario monitoring, the RSBB provides a single interface for smart home features through which the robot can interact with devices in the home. Importantly, the virtual referee employs the same ROS topics used in the RSBB in the physical competition, meaning teams don't have to change their solution to interact with our referee. Once the user launches the referee, a start signal is sent to the robot which contains the scenario context. The scenario context could be: 

\begin{itemize}
    \item The smart doorbell has been activated and the robot must greet the visitor.
    \item Granny Annie has used her smart tablet to ``summon" the robot to her.
    \item Granny Annie makes a natural language request.
\end{itemize}

The virtual referee then waits for the robot to signal it has completed the task. Once the signal is received, the state of the environment (using gazebo model states) is examined to provide a score for the run.

The virtual benchmark contains the following \textit{scenario}, a simplified version of the ERL Consumer competition task ``catering for Granny Annie's comfort". Granny Annie summons the robot through the smart tablet and requests the cracker box from the kitchen. The robot must interpret this command and retrieve the box. Once the task is completed, the robot notifies the referee and receives a score of two for (1) removing the box from the kitchen island and (2) placing the box next to Granny Annie.

% could explain why this is simpler if struggling for words

%\subsection{Extending the virtual benchmark}
% describe the process for adding 

The virtual benchmark is designed to be extended with additional test scenarios, which requires two steps to set up. Firstly, the user must configure the world state by producing a new \texttt{roslaunch} file to describe the objects and starting location for the robot. Secondly, the user must add the scenario logic to the virtual referee. This can be accomplished by extending the existing referee with the start signal and end logic, reusing the topics and services already in place.

The virtual benchmark does not rely on a particular robot or software solution to be compatible. The only requirement is the ROS interface, which is also true for the physical competition. Currently the virtual benchmark is configured for the RALT environment and Tiago robot. However, the \texttt{roslaunch} file can be reconfigured to use another environment/robot and reuse the virtual referee scenario logic freely. This means that it is possible to test your solution across multiple competition testbeds to assess generalisation capability\cite{ISRtestbed}. 

\section{Conclusion and Further Work}
% Research limitations and implications 
Implementing automatic scoring benchmarks provides a potential tool for holding robotics competitions virtually. They can also be applied to other similar competitions such as RoboCup@Home. The tasks and functional benchmarks provide obvious candidates for developing further scenarios, but we can also create new situations to bridge the gap in complexity between benchmark and task. Finally, incorporating elements of noise (e.g. human models active in the environment) and uncertainty (e.g. sensor readings)  can help improve generalisability through making the virtual testbed more similar to a real world environment.

Looking beyond COVID-19, virtual benchmarks provide milestones and test cases for team's systems, thus providing a means for developers to test their system without physical access. They can help improve the generalisation of a system by making it easy to test across many virtual benchmarks. Finally, we believe automation of the environment set up will allow teams to spend more time focusing on their strategy, ultimately leading to more innovative systems.

% Don't say.. Visual components will be harder to test in the virtual environment, this is the well known problem of the reality gap. 

\bibliographystyle{IEEEtran}
\bibliography{references}

\end{document}